%% file: acl2023.tex
\pdfoutput=1

\documentclass[11pt]{article}

\usepackage[final]{ACL2023}
\usepackage{enumitem} 

\usepackage{times}
\usepackage{latexsym}
\usepackage{colortbl}

\usepackage[T1]{fontenc}

\usepackage[utf8]{inputenc}

\usepackage{microtype}
\usepackage{hyperref}

\usepackage{inconsolata}

\usepackage{graphicx}

\usepackage{booktabs}
\usepackage[intlimits]{amsmath}  
\usepackage[colorinlistoftodos,size=tiny]{todonotes}
\usepackage{caption}
\usepackage{pifont}
\usepackage{subcaption}
\usepackage{multirow}
\usepackage{enumitem}
\usepackage{makecell}
\usepackage{tabularx}
\usepackage{soul}
\usepackage{amssymb}

\usepackage{xcolor}
\usepackage{tcolorbox}

\usepackage{booktabs}
\usepackage{tabularx}

\definecolor{criticalred}{RGB}{220,50,50}
\definecolor{majororange}{RGB}{230,150,20}
\definecolor{minorblue}{RGB}{30,100,200}

\tcbset{
    reasoning/.style={
        enhanced,
        colback=gray!5,
        colframe=gray!60,
        fonttitle=\bfseries,
        boxrule=1pt,
        boxsep=6pt,
        arc=3pt,
        breakable
    }
}

\title{DeepSeek-R1 vs.\ o3-mini: \\ How Well can Reasoning LLMs Evaluate MT and Summarization?}

\author{
    Daniil Larionov$^\dagger$,
    Sotaro Takeshita$^\dagger$,
    Ran Zhang$^\dagger$,
    Yanran Chen$^\diamond$\\
    \textbf{Christoph Leiter$^\dagger$,
    Zhipin Wang$^\diamond$,
    Christian Greisinger$^\diamond$,
    Steffen Eger$^\diamond$}\\
    Natural Language Learning \& Generation (NLLG) \url{https://nl2g.github.io/} \\
    $^\dagger$University of Mannheim, $^\diamond$University of Technology Nuremberg
    \\
    \texttt{
        daniil.larionov@uni-mannheim.de
    }
}

\begin{document}
\maketitle
\begin{abstract}
Reasoning-enabled large language models (LLMs) excel in logical tasks, yet their utility for evaluating natural language generation remains unexplored. This study systematically compares reasoning LLMs with non-reasoning counterparts across machine translation and text summarization evaluation tasks. We evaluate eight models spanning state-of-the-art reasoning models (DeepSeek-R1, OpenAI o3), their distilled variants (8B-70B parameters), and equivalent non-reasoning LLMs. Experiments on WMT23 and SummEval benchmarks reveal architecture and task-dependent benefits: OpenAI o3-mini models show improved performance with increased reasoning on MT, while DeepSeek-R1 and generally underperforms compared to its non-reasoning variant except in summarization consistency evaluation. Correlation analysis demonstrates that reasoning token usage correlates with evaluation quality only in specific models, while almost all models generally allocate more reasoning tokens when identifying more quality issues. Distillation maintains reasonable performance up to 32B parameter models but degrades substantially at 8B scale. This work provides {\bf the first assessment of reasoning LLMs for NLG evaluation} and comparison to non-reasoning models. We share our code to facilitate further research: \url{https://github.com/NL2G/reasoning-eval}.
\end{abstract}


\input{structure/1_introduction}
\input{structure/2_related_work}
\input{structure/3_experimental_setup}
\input{structure/4_summarization_results}
\input{structure/5_mt_results}
\input{structure/6_analysis}
\input{structure/7_discussion}

\input{structure/8_conclusion}

\section*{Limitations}
There are several limitations associated with the evaluation methodology we used. Future work may address them.
\begin{enumerate}
\item We make an assumption that the GPT-4o-mini model is the most likely non-reasoning equivalent, by capabilities and/or size, to the reasoning model o3-mini. At the same time, there is no public confirmed information about the size of either of these models, thus this cannot be verified.
\item While the inference API of o3-mini allows specifying the reasoning effort (either `low', `medium' or `high'), there is no such option for DeepSeek R1 and its distilled variants. Further studies might examine R1 performance at different reasoning efforts by implementing forced generation of the closing `</think>' token upon generating a pre-determined amount of reasoning tokens.
\item We have not touched the Hybrid Reasoning LLMs, such as Anthropic Claude 3.7 or most recently released Qwen3. The hybrid approach in those models allows to turn on or off the generation of reasoning trace before providing the result via a special command tokens. Those models could potentially exhibit smaller discrepancy in quality between reasoning/non-reasoning, as the post-training phase of those models is unified.
\end{enumerate}

\section*{Acknowledgements}
\begin{figure*}
    \centering
    \includegraphics[width=0.45\linewidth]{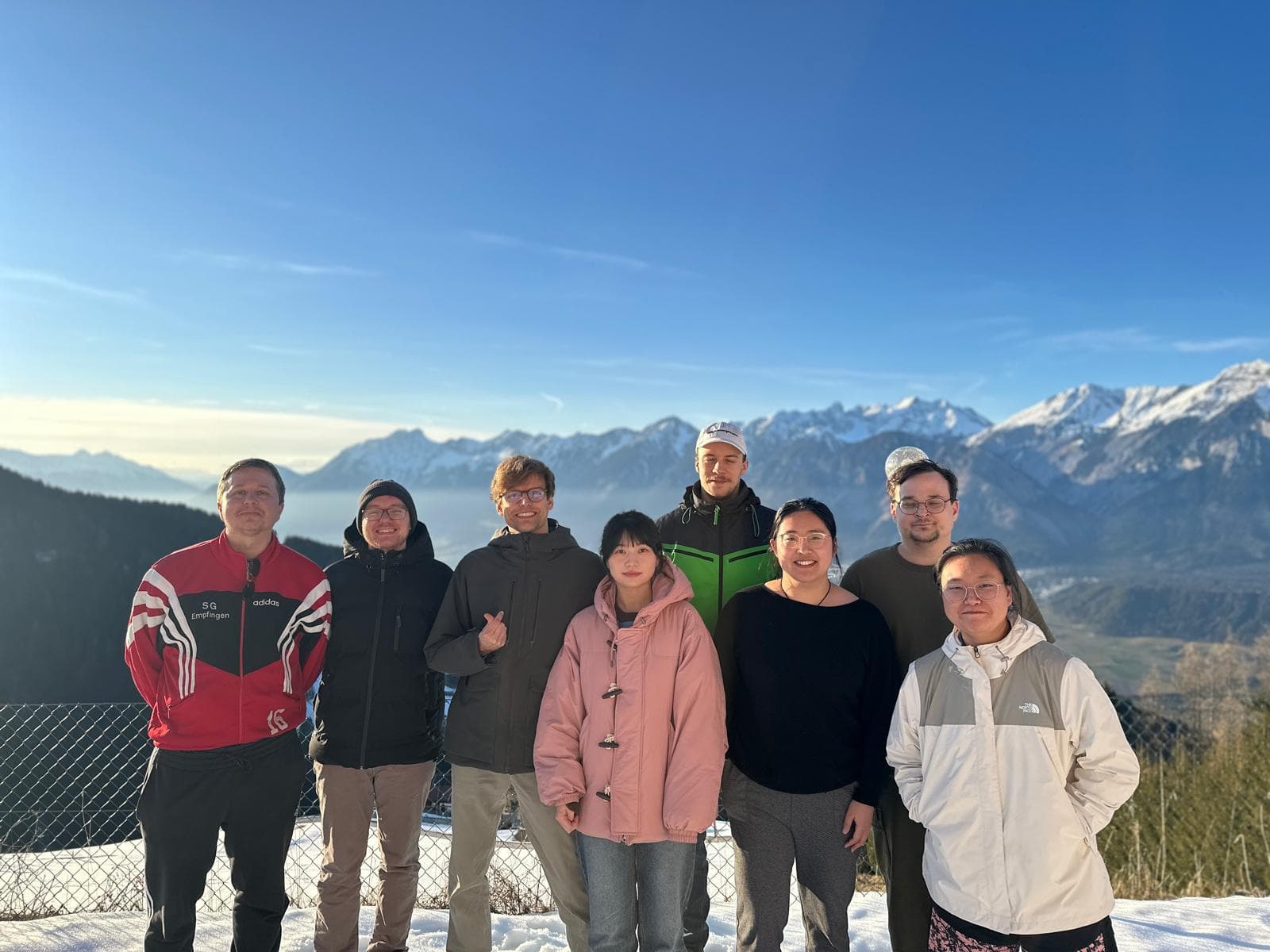}
    \includegraphics[width=0.45\linewidth]{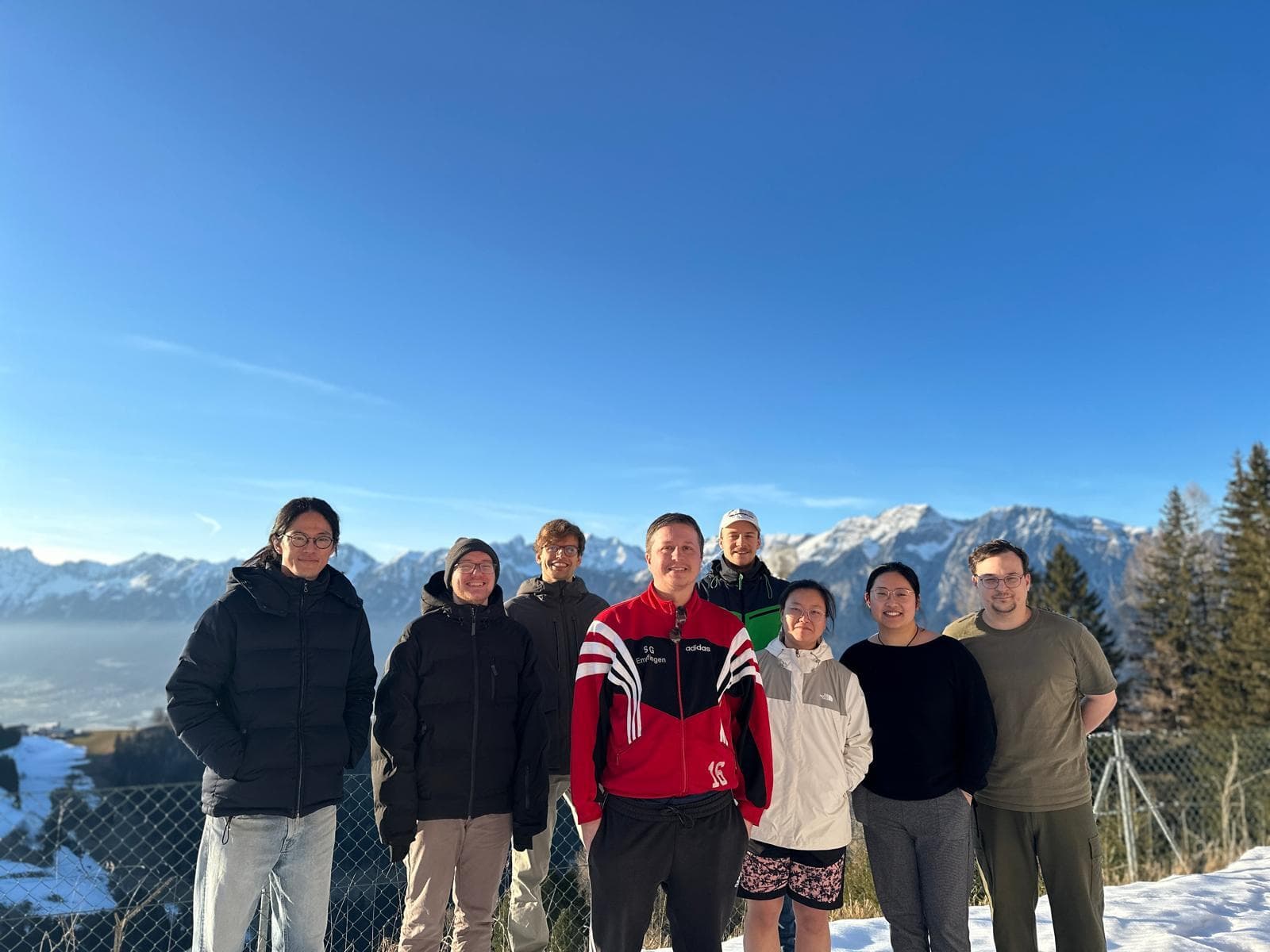}
    \caption{The NLLG lab in Großvolderberg during February, 2025.}
    \label{fig:retreat}
\end{figure*}

The NLLG lab gratefully acknowledges support from the Federal Ministry of Education and Research (BMBF) via the research grant ``Metrics4NLG'' and the German Research Foundation (DFG) via the Heisenberg Grant EG 375/5–1. Parts of this paper were developed during a retreat in the Tyrolean Alps, in the settlement of Großvolderberg. See Figure \ref{fig:retreat}.

\bibliography{anthology,custom}
\bibliographystyle{acl_natbib}

\appendix

\input{structure/X-Appendix}

%

\end{document}

%% file: structure/1_introduction.tex
\section{Introduction}

Reasoning LLMs have driven  recent progress in NLP, often outperforming standard models by leveraging chain-of-thought (CoT) supervision~\citep{10.5555/3600270.3602070} and reinforcement learning~\citep{shao2024deepseekmathpushinglimitsmathematical} to improve multi-step and logical inference~\citep{arora2025traininglanguagemodelsreason, guo2025deepseek}. 

\begin{figure}[htbp]
    \centering
    \includegraphics[width=1\linewidth]{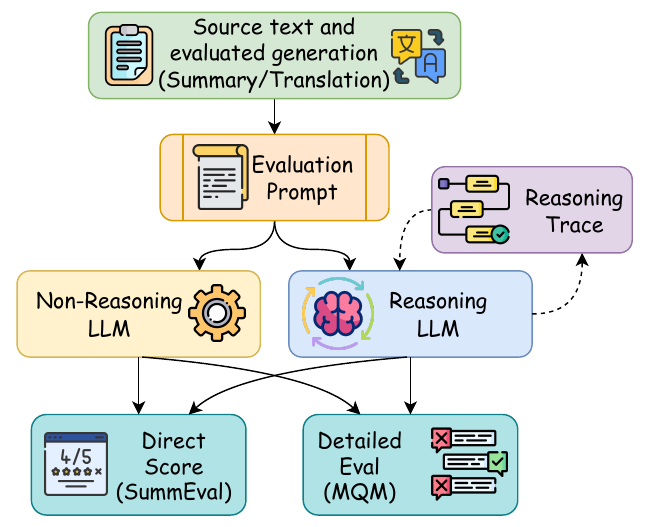}
    \caption{Reasoning and Non-Reasoning LLMs for Prompt-Based Evaluation.}
    \label{fig:reasoning}
\end{figure}

However, reasoning is not always beneficial---CoT prompting can degrade performance when verbal deliberation introduces unnecessary complexity~\citep{liu2024mind} or the task does not require it~\citep{sprague2025to}. This shows the need to explore when reasoning is actually advantageous.

A critical but underexplored application is Natural Language Generation (NLG) evaluation. Although LLM-based metrics have become the go-to approach for high-quality machine translation (MT) and text summarization (TS) evaluation~\citep{kocmi-federmann-2023-gemba, liu-etal-2023-g, leiter-etal-2023-eval4nlp}, existing methods are primarily built on non-reasoning LLMs. Reasoning could improve NLG evaluation by enabling nuanced assessment processes that mirror how human experts evaluate translations and summaries, that is, through step-by-step analysis of semantic preservation, identification of subtle contextual mismatches, and systematic weighing of multiple quality dimensions that often requires multi-step thinking rather than holistic pattern matching. This gap raises two key research questions: (1) whether reasoning models do improve NLG evaluation quality, and (2) whether distilled reasoning models can maintain evaluation capabilities while reducing computational costs, given that full-scale reasoning LLMs often have several hundreds of billions parameters. 

To address these questions, we systematically test reasoning LLMs and compare them to equivalent non-reasoning models. The overall prompt-based evaluation pipeline is illustrated in Figure~\ref{fig:reasoning}. Our framework includes three architectural categories: (1) SOTA reasoning LLMs: DeepSeek-R1 and OpenAI o3; (2) distilled variants of R1 with 70, 32 and 8 billion parameters; and (3) equivalent non-reasoning LLMs, such as GPT-4o-mini, Qwen-2.5 32B, and Llama-3.1 8B. We test these models using G-Eval~\citep{liu-etal-2023-g} for TS and GEMBA-MQM~\citep{kocmi-federmann-2023-gemba} for MT evaluation.

Our results reveal distinct performance patterns across model families: {\bf a)} The OpenAI o3-mini model outperforms non-reasoning baselines on MT evaluation metrics and delivers substantial improvements on `coherence' and `fluency' dimensions in TS evaluation; {\bf b)} DeepSeek-R1 and its distilled versions underperform their non-reasoning counterparts on most MT metrics, with some exceptions in TS `consistency' and `fluency' quality aspects; {\bf c)} Correlation analysis identifies a low-medium correlation between reasoning token count and evaluation quality for the OpenAI o3-mini for MT evaluation and R1-Distill-Qwen-32B for TS evaluation, but very low to no correlation for other models in the study; {\bf d)} Models consistently dedicate more reasoning tokens when assessing texts that they find low-quality compared to high-quality ones, with a strong negative correlation between reasoning length and predicted quality score; {\bf e)} Distillation up to 32B parameters preserves most evaluation quality for reasoning models while 8B models exhibit substantial performance degradation on the studied evaluation tasks.

Our work presents the first systematic comparison of reasoning versus non-reasoning LLMs as NLG evaluators and provides valuable insights into the efficacy of reasoning distillation for practical deployment in NLG evaluation tasks.

%% file: structure/2_related_work.tex
\section{Related Work}

Our work connects to evaluation metrics for MT, summarization and ``LLM-as-a-Judge'' approaches, efficiency in  metrics as well as to reasoning in LLMs.

\paragraph{Traditional Metrics for MT}
Early MT evaluation metrics focused on surface-level similarity between system and reference translations. BLEU~\citep{papineni-etal-2002-bleu} and METEOR~\citep{banerjee-lavie-2005-meteor} rely on n-gram overlap and edit distance. BERTScore~\citep{Zhang*2020BERTScore:} further improved MT evaluation by computing contextualized word embeddings from BERT~\citep{devlin-etal-2019-bert}. To capture semantic shifts more effectively, MoverScore~\citep{zhao-etal-2019-moverscore} leverages word embeddings to measure distances between system and reference translations, using Word Mover’s Distance~\citep{10.5555/3045118.3045221}. Other paradigms popular for evaluating MT include NLI-based metrics \citep{chen-eger-2023-menli} and trained metrics \citep{rei-etal-2020-comet,guerreiro-etal-2024-xcomet}.
\paragraph{Traditional Metrics for TS} 
One of the foundational metrics for evaluating TS is ROUGE~\citep{lin-2004-rouge}, which measures the overlap between system-generated summaries and reference summaries. However, ROUGE struggles with semantic equivalence and factual consistency, particularly in long-form summarization. To address these limitations, content-based metrics such as Pyramid~\citep{nenkova-passonneau-2004-evaluating} and BE~\citep{hovy-etal-2006-automated} were introduced and later semantic metrics such as SUPERT~\citep{gao-etal-2020-supert}. Pyramid aligns important information across multiple references, while BE identifies minimal semantic units in a reference summary. Benchmarking efforts such as SummEval~\citep{fabbri-etal-2021-summeval} have further supported systematic comparisons of TS evaluation metrics.

\paragraph{LLM-based Evaluation}

The ``LLM-as-a-Judge'' paradigm~\citep{gu2024survey, li2025generationjudgmentopportunitieschallenges, huang2024empiricalstudyllmasajudgellm} employs instruction-following LLMs (e.g., GPT-4~\citep{openai2024gpt4ocard}, LLaMA-3~\citep{grattafiori2024llama3herdmodels}) to directly assess system outputs along dimensions such as fluency, coherence, and factual consistency~\citep{chia2023instructeval}. These methods increasingly correlate with human judgment and offer a scalable alternative to traditional metrics~\citep{liu-etal-2023-g, kocmi-federmann-2023-gemba}. In MT, GEMBA~\citep{kocmi-federmann-2023-gemba, kocmi-federmann-2023-large} prompts LLMs to evaluate individual segments in isolation before aggregating scores, achieving state-of-the-art performance on the WMT22 Metrics Benchmark~\citep{freitag-etal-2022-results}. AutoMQM~\citep{DBLP:conf/wmt/FernandesDFRMNG23} extends this by asking LLMs to detect and classify translation errors, offering more interpretable and fine-grained feedback. In TS, G-Eval~\citep{liu-etal-2023-g} adopts a criteria-based framework using chain-of-thought prompting, while FineSurE~\citep{song-etal-2024-finesure} incorporates key fact extraction and verification by LLMs. Eval4NLP~\citep{leiter-etal-2023-eval4nlp} further benchmarked prompting-based LLM evaluators for summarization in a shared task setting. Despite these advances, existing work leverages only non-reasoning LLMs. The recent release of reasoning-enabled LLMs opens the question of whether reasoning LLMs offer further benefits for the `LLM-as-a-Judge' paradigm.

\paragraph{Efficiency for Evaluation Metrics}

With the rising computational cost of LLM-based evaluation, recent work has explored lightweight alternatives to maintain effectiveness under resource constraints. FrugalScore~\citep{eddine2022frugalscore} uses distillation and representation pruning to reduce model size without compromising metric reliability. EffEval~\citep{larionov-etal-2023-effeval} investigates trade-offs between efficiency and quality, showing that adapter-based fine-tuning and compact architectures are viable. Similarly, COMETINHO~\citep{rei-etal-2022-searching} introduces efficient variants of COMET through optimized inference, while xCOMET-lite~\citep{larionov-etal-2024-xcomet} leverages quantization and distillation for real-time applications. Prompt-level optimization methods such as PromptOptMe~\citep{larionov-eger-2025-promptoptme} reduce token overhead while preserving evaluation accuracy. Building on previous efforts in efficient evaluation, we investigate how effective  smaller,  distilled reasoning models are at NLG evaluation.

\paragraph{Reasoning LLMs}
Recent models such as DeepSeek-R1~\citep{guo2025deepseek} and Open AI o3-mini\footnote{\url{https://openai.com/index/openai-o3-mini/}} are designed to perform complex step-by-step reasoning using techniques like CoT prompting~\citep{10.5555/3600270.3602070} through Group Relative Policy Optimization (GRPO)~\citep{shao2024deepseekmathpushinglimitsmathematical}. These approaches enable intermediate reasoning supervision and reward shaping~\citep{arora2025traininglanguagemodelsreason}, leading to strong performance on mathematically and logically demanding tasks. Further, efficiency and quality gains are achieved through CoT compression~\citep{deng2024explicitcotimplicitcot}, model distillation~\citep{guo2025deepseek}, and test-time scaling strategies~\citep{muennighoff2025s1simpletesttimescaling}. However, their potential for evaluating NLG remains underexplored, as existing approaches often use heuristic CoT prompting without fully leveraging reasoning LLMs \citep{leiter-eger-2024-prexme}. This motivates our investigation into whether such models can outperform conventional LLMs.

%% file: structure/3_experimental_setup.tex
\section{Experimental Setup}

Our evaluation focuses on two primary NLG tasks: MT evaluation and TS evaluation. We choose MT evaluation as it allows to asses multilingual evaluation capabilities of reasoning LLMs. Summarization, in turn, was chosen as it allows us to evaluation longer-context capabilities of LLMs, because it requires analyzing long original texts and their summaries. For this, the reasoning might potentially be especially helpful. We utilize established prompting-based metrics that have demonstrated strong performance in recent benchmarks.

\paragraph{MT Evaluation} For MT evaluation, we employ GEMBA-MQM~\citep{kocmi-federmann-2023-gemba}, a prompting-based metric that leverages LLMs to identify and rate translation quality errors. GEMBA-MQM instructs LLMs to evaluate MT by detecting errors across dimensions such as accuracy, fluency, and terminology, assigning severity scores to each error. This method has shown state-of-the-art performance on the WMT22 Metrics Shared Task~\citep{freitag-etal-2023-results}.

\paragraph{TS Evaluation} For TS evaluation, we implement G-Eval~\citep{liu-etal-2023-g}, which prompts LLMs to evaluate summaries along multiple dimensions including coherence, consistency, relevance, and fluency. G-Eval provides explicit criteria for each dimension and asks the model to generate both numerical scores and explanations, making it particularly suitable for testing the impact of reasoning capabilities.

\paragraph{Datasets}
For MT we utilize the WMT23 Metrics Shared Task dataset~\citep{freitag-etal-2023-results}, which provides human judgments for translations between various language pairs. The dataset includes source texts, reference translations, system translations, and human judgments at segment levels. We use the en-de, he-en, and zh-en language pairs to evaluate across diverse linguistic structures. Each translation is accompanied by span-level error annotations.

For TS evaluation, we use the SummEval dataset~\citep{fabbri-etal-2021-summeval}, which contains human judgments for summaries from 16 different summarization systems on 100 news articles from the CNN/DailyMail dataset. Each summary is evaluated along four dimensions: coherence, consistency, relevance, and fluency, with scores ranging from 1 to 5. Additionally, we employ the Eval4NLP dataset~\citep{leiter-etal-2023-eval4nlp}. The human judgement scores for Eval4NLP are not published online, which allows us to evaluate reasoning and non-reasoning models with no data contamination.

\paragraph{Models}
We evaluate three categories of models to systematically investigate our research questions:

\paragraph{SOTA Reasoning LLMs:} \textbf{DeepSeek R1}~\citep{guo2025deepseek} is a reasoning-enabled variant of DeepSeek V3, fine-tuned using reinforcement learning for explicit generation of reasoning. This 600B parameter model represents the current frontier of reasoning architecture. \textbf{OpenAI o3-mini} is a reasoning model with controllable reasoning effort: high, medium, and low. This model allows us to investigate the impact of reasoning intensity on evaluation performance within the same model architecture.

\paragraph{Distilled Variants of DeepSeek R1:} \textbf{R1 LLaMA 70B}, \textbf{R1 Qwen 32B}, and \textbf{R1 LLaMA 8B}, representing three different LLM size tiers.

\paragraph{Non-Reasoning LLMs (Control Group):} \textbf{DeepSeek V3}~\citep{deepseekai2025deepseekv3technicalreport} as non-reasoning counterpart to DeepSeek R1; \textbf{GPT-4o-mini}~\citep{openai2024gpt4ocard} as presumed non-reasoning variant from the GPT-4 family; \textbf{Qwen-2.5 32B}~\citep{qwen2025qwen25technicalreport} as counterpart for R1 Distill Qwen 32B; \textbf{LLaMa-3.3 70B}~\citep{grattafiori2024llama3herdmodels} as counterpart to R1 Distill Llama 70B and \textbf{LLaMa-3.1 8B}~\citep{grattafiori2024llama3herdmodels} for R1 LLaMa 8B.

This selection provides a comprehensive cross-section of reasoning vs.\ non-reasoning models across different parameter scales, allowing us to analyze both capability differences and the impact of distillation.

\paragraph{Evaluation Protocol}

All models were evaluated using the same prompting templates to ensure fair comparison. For the GEMBA-MQM metric, we followed the template described in~\citet{kocmi-federmann-2023-gemba}, 
which instructs the model to identify errors in a translation and then score the translation based on those errors. For G-Eval, we used the prompts detailed in~\citet{liu-etal-2023-g}, 
which instructs the model to evaluate summaries on multiple dimensions with specific criteria.

For meta-evaluation of MT metrics, we compute segment-level Pearson correlation ($\rho$) between model scores and human judgments, as well as system-level pairwise accuracy~\citep{deutsch-etal-2023-ties}, which measures how often the metric correctly predicts which of two systems is better according to human judgments. For TS evaluation, we compute segment-level Kendall $\tau$ correlation with human judgments across quality aspects.

%% file: structure/4_summarization_results.tex
\section{Results}
Here, we describe the results obtained with TS evaluation and MT evaluation.
\subsection{Summarization}
\input{structure/tables/combined}
Table \ref{tab:combined_eval} shows the segment-level Kendall's $\tau$ correlation between human judgments and model evaluations on the SummEval and Eval4NLP datasets. For SummEval, we compare performance across four dimensions: coherence, consistency, relevance, and fluency. For Eval4NLP, we report correlation with a single summarization quality score.
\paragraph{OpenAI vs.\ DeepSeek}
OpenAI reasoning models show consistent performance irrespective of the reasoning effort setting, with o3-mini-high and o3-mini-low achieving nearly identical SummEval averages (0.337 and 0.335) and the highest overall Eval4NLP scores (0.644 and 0.645). DeepSeek V3 achieves the highest SummEval average score (0.399) among all models tested, with particular strengths in coherence (0.462), relevance (0.446), and fluency (0.356).
\paragraph{Reasoning vs.\ non-reasoning models}
Across model families, we identify a pattern of non-reasoning models outperforming their reasoning counterparts: {\bf 1)} DeepSeek V3 (non-reasoning) outperforms DeepSeek R1 in both SummEval average (0.399 vs.\ 0.351) and Eval4NLP (0.630 vs.\ 0.583); {\bf 2)} LLaMA 3.3 70B (non-reasoning) achieves better scores than R1 LLaMa 70B in both SummEval average (0.375 vs.\ 0.315) and Eval4NLP (0.624 vs.\ 0.556); {\bf 3)} Qwen 2.5 32B (non-reasoning) outperforms R1 Qwen 32B with higher SummEval average (0.393 vs.\ 0.355) and Eval4NLP (0.619 vs.\ 0.564) scores; and {\bf 4)} The OpenAI reasoning and non-reasoning models achieve more comparable scores for SummEval average (0.337 vs.\ 0.346) and Eval4NLP (0.644 vs.\ 0.634).

However, reasoning models demonstrate specific strengths in evaluation of consistency, coherence, and fluency. DeepSeek R1 excels in consistency metrics (0.565 vs.\ 0.331, a 70\% better correlation), showing substantially better alignment with human judgments as compared to the non-reasoning DeepSeek V3. Similarly, R1 Qwen 32B shows strong consistency performance, outperforming Qwen 2.5 32B by 20\% (0.540 vs.\ 0.449). R1 LLaMa 70B in turn outperforms its non-reasoning counterpart in fluency evaluation by 16\% (0.330 vs.\ 0.284). The OpenAI o3-mini model outperforms the assumed non-reasoning equivalent GPT-4o-mini in coherence by approximately 50\% (0.482 vs.\ 0.321) and in fluency by 25\% (0.313 vs.\ 0.263).
\paragraph{Distilled vs.\ original models}
When comparing the original DeepSeek R1 with its distilled variants, we observe that R1 Qwen 32B matches the original model quality on SummEval average and achieves 97\% on Eval4NLP (0.564 vs.\ 0.583). This distilled model has 21 times fewer parameters than the original one (32B vs.\ 685B). The distillation process also largely maintains the reasoning model advantage in consistency evaluation over equivalent non-reasoning models.

At the same time, the small distilled model (R1 LLaMa 8B) shows substantially worse performance compared to the original R1 model on the SummEval dataset. We can also see that it loses the advantage in the consistency aspect of TS evaluation.

%% file: structure/tables/combined.tex
\begin{table*}[htbp]\centering
\resizebox{\linewidth}{!}{
\begin{tabular}{llcccccc|cccc}\toprule
\multirow{2}{*}{\textbf{Model Name}} &\multirow{2}{*}{\textbf{Reas.}} &\multicolumn{5}{c}{\textbf{SummEval}} &\textbf{Eval4NLP} &\multicolumn{3}{c}{\textbf{WMT23 seg-level}} &\textbf{sys-level} \\\cmidrule(lr){3-7}\cmidrule(lr){8-8}\cmidrule(lr){9-11}\cmidrule(lr){12-12}
& &\textbf{coh} &\textbf{con} &\textbf{rel} &\textbf{flu} &\textbf{avg.} & &\textbf{en-de} &\textbf{he-en} &\textbf{zh-en} & \textbf{acc.}\\\midrule
\multicolumn{12}{c}{\textbf{DeepSeek, Qwen, LLaMa}} \\\midrule
\textbf{DeepSeek R1} &yes &\cellcolor[HTML]{99d6b8}0.381 &\cellcolor[HTML]{57bb8a}0.565 &\cellcolor[HTML]{b5e1cb}0.303 &\cellcolor[HTML]{e9f6f0}0.157 &\cellcolor[HTML]{a4dabf}0.351 &\cellcolor[HTML]{ffe089}0.583 &\cellcolor[HTML]{a8dcc2}0.364 &\cellcolor[HTML]{95d4b5}0.398 &\cellcolor[HTML]{88cfac}0.441 &\cellcolor[HTML]{ffdb78}0.908\\
\textbf{DeepSeek V3} &no &\cellcolor[HTML]{7ccaa4}0.462 &\cellcolor[HTML]{abddc5}0.331 &\cellcolor[HTML]{82cda8}0.446 &\cellcolor[HTML]{a2dabe}0.356 &\cellcolor[HTML]{93d3b4}0.399 &\cellcolor[HTML]{ffd96f}0.630 &\cellcolor[HTML]{79c9a2}0.490 &\cellcolor[HTML]{96d5b6}0.394 &\cellcolor[HTML]{74c89f}0.512 &\cellcolor[HTML]{ffdb7a}0.904\\
\textbf{R1 LLaMa 70B} &yes &\cellcolor[HTML]{99d6b8}0.380 &\cellcolor[HTML]{d0ecde}0.228 &\cellcolor[HTML]{aedfc7}0.322 &\cellcolor[HTML]{abddc5}0.330 &\cellcolor[HTML]{b1e0c9}0.315 &\cellcolor[HTML]{ffe498}0.556 &\cellcolor[HTML]{90d1b1}0.421 &\cellcolor[HTML]{a5dbc0}0.365 &\cellcolor[HTML]{86ceab}0.451 &\cellcolor[HTML]{ffd86d}0.932\\
\textbf{LLaMa 3.3 70B} &no &\cellcolor[HTML]{73c79e}0.487 &\cellcolor[HTML]{b8e3ce}0.293 &\cellcolor[HTML]{85ceaa}0.437 &\cellcolor[HTML]{bce4d0}0.284 &\cellcolor[HTML]{9bd7b9}0.375 &\cellcolor[HTML]{ffda72}0.624 &\cellcolor[HTML]{60bf91}0.590 &\cellcolor[HTML]{90d1b1}0.420 &\cellcolor[HTML]{6ec599}0.522 &\cellcolor[HTML]{ffda72}0.924\\
\textbf{R1 Qwen 32B} &yes &\cellcolor[HTML]{b8e3ce}0.293 &\cellcolor[HTML]{60bf91}0.540 &\cellcolor[HTML]{b4e1cb}0.304 &\cellcolor[HTML]{bde4d1}0.281 &\cellcolor[HTML]{a2dabf}0.355 &\cellcolor[HTML]{ffe394}0.564 &\cellcolor[HTML]{9bd7b9}0.388 &\cellcolor[HTML]{aedfc7}0.338 &\cellcolor[HTML]{81cca7}0.465 &\cellcolor[HTML]{ffdb78}0.920\\
\textbf{Qwen2.5 32B} &no &\cellcolor[HTML]{9cd7ba}0.372 &\cellcolor[HTML]{81cca7}0.449 &\cellcolor[HTML]{91d3b2}0.404 &\cellcolor[HTML]{a5dbc0}0.348 &\cellcolor[HTML]{95d4b5}0.393 &\cellcolor[HTML]{ffda75}0.619 &\cellcolor[HTML]{6fc599}0.521 &\cellcolor[HTML]{98d6b8}0.390 &\cellcolor[HTML]{70c69a}0.519 &\cellcolor[HTML]{ffd666}0.944\\
\textbf{R1 LLaMa 8B} &yes &\cellcolor[HTML]{e5f5ed}0.169 &\cellcolor[HTML]{c7e9d8}0.251 &\cellcolor[HTML]{f3fbf7}0.128 &\cellcolor[HTML]{ecf8f2}0.148 &\cellcolor[HTML]{e3f4ec}0.174 &0.368 &\cellcolor[HTML]{a0d9bc}0.310 &\cellcolor[HTML]{b2e1c9}0.325 &\cellcolor[HTML]{8fd1b0}0.410 &\cellcolor[HTML]{ffdc7c}0.915\\
\textbf{LLaMa 3.1 8B} &no &\cellcolor[HTML]{eef9f4}0.142 &\cellcolor[HTML]{b2e0c9}0.312 &\cellcolor[HTML]{9fd9bc}0.364 &0.094 &\cellcolor[HTML]{d0ecde}0.228 &\cellcolor[HTML]{ffeebe}0.488 &\cellcolor[HTML]{7dcba5}0.476 &\cellcolor[HTML]{aedfc7}0.335 &\cellcolor[HTML]{90d1b1}0.421 &\cellcolor[HTML]{ffdc7d}0.916\\
\multicolumn{12}{c}{\textbf{OpenAI}} \\\midrule
\textbf{o3-mini-high} &yes &\cellcolor[HTML]{75c79f}0.482 &\cellcolor[HTML]{cbeadb}0.242 &\cellcolor[HTML]{b2e0ca}0.311 &\cellcolor[HTML]{b2e0c9}0.311 &\cellcolor[HTML]{a9dcc3}0.337 &\cellcolor[HTML]{ffd767}0.644 &\cellcolor[HTML]{62c093}0.577 &\cellcolor[HTML]{90d1b1}0.421 &\cellcolor[HTML]{62c093}0.568 &\cellcolor[HTML]{ffdb78}0.920\\
\textbf{o3-mini-medium} &yes &\cellcolor[HTML]{75c7a0}0.479 &\cellcolor[HTML]{cdeadc}0.235 &\cellcolor[HTML]{b2e0ca}0.309 &\cellcolor[HTML]{b1e0c9}0.311 &\cellcolor[HTML]{a9ddc3}0.333 &\cellcolor[HTML]{ffd767}0.644 &\cellcolor[HTML]{71c69b}0.517 &\cellcolor[HTML]{93d3b4}0.404 &\cellcolor[HTML]{76c8a0}0.505 &\cellcolor[HTML]{ffd96f}0.928\\
\textbf{o3-mini-low} &yes &\cellcolor[HTML]{76c8a0}0.478 &\cellcolor[HTML]{cdebdc}0.237 &\cellcolor[HTML]{b1e0c9}0.315 &\cellcolor[HTML]{b1e0c9}0.313 &\cellcolor[HTML]{a9ddc3}0.335 &\cellcolor[HTML]{ffd666}0.645 &\cellcolor[HTML]{7fcba6}0.471 &\cellcolor[HTML]{91d2b2}0.413 &\cellcolor[HTML]{7bc9a3}0.491 &\cellcolor[HTML]{ffd96f}0.928\\
\textbf{GPT-4o-mini} &no &\cellcolor[HTML]{aedfc7}0.321 &\cellcolor[HTML]{88cfac}0.430 &\cellcolor[HTML]{9dd8bb}0.370 &\cellcolor[HTML]{c3e7d6}0.263 &\cellcolor[HTML]{a6dbc1}0.346 &\cellcolor[HTML]{ffd86d}0.634 &\cellcolor[HTML]{93d3b4}0.410 &\cellcolor[HTML]{8ad0ad}0.435 &\cellcolor[HTML]{7cc9a3}0.487 &\cellcolor[HTML]{ffd96f}0.928\\
\bottomrule
\end{tabular}
}
\caption{Combined results for TS and MT evaluation across models. For TS evaluation we show segment-level Kendall's $\tau$ on SummEval across quality aspects (coherence, consistency, relevance, fluency), average, and correlation for Eval4NLP dataset. For MT evaluation we provide segment-level Pearson's $\rho$ (en-de, he-en, zh-en) and system-level pairwise accuracy on WMT23. `Reas.' indicates if the LLM is reasoning-enabled or not.}
\label{tab:combined_eval}
\end{table*}

%% file: structure/5_mt_results.tex
\subsection{Machine Translation}

Table~\ref{tab:combined_eval} shows the segment-level Pearson's $\rho$ correlation between human judgments and model evaluations on the WMT23 dataset across three language pairs: en-de, he-en, and zh-en. Additionally, we report system-level pairwise accuracy, measuring how often the metric correctly ranks translation systems. All measures were computed using `mt-metrics-eval' package~\footnote{\url{https://github.com/google-research/mt-metrics-eval}}. 

\paragraph{OpenAI vs.\ DeepSeek} The OpenAI models demonstrate strong performance across language pairs, with o3-mini-high achieving the highest overall segment-level correlations (0.577 for en-de and 0.568 for zh-en). Within the OpenAI o3-mini variants, we observe that higher reasoning effort settings generally correspond to better performance, particularly for the en-de and zh-en pairs. Interestingly, GPT-4o-mini performs comparatively well on low-resource language pair he-en (0.435), outperforming its reasoning counterparts. For system-level accuracy, the OpenAI models achieve identical scores (0.928 for o3-mini variants and GPT-4o-mini). 

\paragraph{Reasoning vs.\ non-reasoning models} Similarly to our summarization findings, certain non-reasoning models  outperform their reasoning counterparts in translation evaluation tasks. Here this pattern holds for DeepSeek models: DeepSeek V3 (non-reasoning) outperforms DeepSeek R1 in both en-de (0.490 vs.\ 0.364) and zh-en (0.512 vs.\  0.441) language pairs, with comparable performance on he-en. R1 LLaMa 70B shows substantial deterioration over LLaMa 3.3 70B across all language pairs, with particularly notable differences in en-de (0.590 vs. 0.421) and he-en (0.420 vs.\ 0.365). Similarly, Qwen 2.5 32B achieves higher correlations than R1 Qwen 32B in all language pairs, with the largest gap in en-de (0.521 vs.\ 0.388).

The exception to this pattern is OpenAI o3-mini at varying reasoning efforts, which demonstrate superior performance compared to GPT-4o-mini across two language pairs: in en-de (0.577 vs.\ 0.410) and zh-en (0.568 vs.\ 0.487).

For system-level accuracy, the differences between reasoning and non-reasoning models are less pronounced. Qwen 2.5 32B achieves the highest system-level accuracy (0.944), with R1 LLaMa 70B (0.932) performing slightly better than its non-reasoning counterpart (0.924).

The distilled reasoning models show varying levels of quality degradation. R1 LLaMa 70B and R1 Qwen2.5 32B outperform original R1 model on two out of three language pairs (en-de and zh-en), while deteriorating on low-resource he-en. At the same time, the smaller R1 LLaMa 8B shows substantial drop across all 3 language pairs.

%% file: structure/6_analysis.tex
\section{Analysis of Reasoning}

To further investigate the relationship between explicit reasoning and evaluation performance, we analyze the correlation between the number of reasoning tokens used by each model and various evaluation metrics. Specifically, we compute Pearson correlations between reasoning token count and: (1) evaluation error (absolute difference between model-predicted and ground truth scores) and (2) model-assigned scores. This analysis was performed for MT evaluation on WMT23 and for TS evaluation on SummEval on all 4 quality aspects. Table~\ref{tab:reasoning-correlations} summarizes these correlations for the reasoning-enabled models. We do not include non-reasoning models here as they naturally do not produce reasoning tokens.

\begin{table}[htbp]
\resizebox{\linewidth}{!}{
\begin{tabular}{lcccc}\toprule
\textbf{Model} & \textbf{MT Error} & \textbf{MT Score} & \textbf{TS Error} & \textbf{TS Score}\\
\midrule
\textbf{DeepSeek R1} & -0.015 & 0.019 & -0.054 & -0.010\\
\textbf{R1 LLaMa 70B} & -0.042 & -0.208 & 0.026 & -0.221\\
\textbf{R1 Qwen 32B} & -0.003$^{\ddagger}$ & -0.150 & -0.122 & -0.216\\
\textbf{o3-mini-high} & -0.118 & -0.474 & -0.156 & -0.209\\
\textbf{o3-mini-medium} & -0.110 & -0.414 & -0.090 & -0.189\\
\textbf{o3-mini-low} & -0.091 & -0.346 & -0.006$^{\ddagger}$ & -0.149\\
\bottomrule
\multicolumn{4}{l}{\scriptsize{$^{\ddagger}$Not statistically significant ($p > 0.05$). All other correlations significant at $p < 0.001$.}}
\end{tabular}
}

\caption{Pearson correlations between reasoning token count and evaluation metrics. {\bf MT} stands for MT evaluation, and {\bf TS} is for TS evaluation. {\bf Error} represents the correlation between reasoning token count and absolute difference between model-predicted and ground truth scores. {\bf Score} is for correlation between between reasoning token count and model-predicted score}\label{tab:reasoning-correlations}
\end{table}

\paragraph{Error Correlations} The majority of reasoning models exhibit statistically significant negative correlations between reasoning token count and evaluation error, suggesting that increased reasoning is associated with smaller discrepancies between model-predicted and human judgment scores. The o3-mini variants show strongest overall negative correlations (-0.118/-0.156, -0.110/-0.090, and -0.091/-0.006 for high, medium, and low effort settings respectively) in MT and TS evaluation, while DeepSeek-based models demonstrate weaker correlations (-0.015/-0.054 for DeepSeek R1, -0.042/0.026 for R1 LLaMa 70B, and -0.003/-0.122 for R1 Qwen 32B). Some models in certain settings exhibit either non-significant correlations (R1 Qwen 32B on MT eval and o3-mini-low on TS eval) or weak positive correlation (R1 LLaMa 70B on TS eval).

\paragraph{LLM Score Correlations} We observe substantial differences in how reasoning token count correlates with model-assigned scores across architectures for both MT and TS evaluation. OpenAI models exhibit strong negative 
correlations for MT and moderate for TS evaluation, with o3-mini-high showing the strongest relationship in MT (-0.4742), followed by o3-mini-medium (-0.4148) and o3-mini-low (-0.3466). The DeepSeek distillations (R1 LLaMa 70B and R1 Qwen 32B) show moderate correlations across MT and TS evaluation. In contrast, original DeepSeek R1 shows a weak correlation for MT and TS eval (0.0199/-0.010). This indicates that OpenAI models (and to some extend DeepSeek distilled models) {\bf tend to engage in a more extensive thinking when they identify more errors} and assign lower quality scores.

\subsection{Examples of Reasoning Traces}
\label{sec:reasoning_analysis}

We present examples of effective and problematic reasoning traces for MT evaluation performed by the DeepSeek R1 model on the En-De language pair in the WMT23 dataset. See Appendix~\ref{sec:appendix_reasoning}.

\begin{figure}[ht]
\small{
\textcolor{purple}{<think>}
Okay, let’s tackle this translation analysis. I compare the English source and German hypothesis to spot errors: “Statistiken Norwegens” misnames “Statistics Norway” \textcolor{magenta}{(critical terminology)}; ... “weiter südlich” vs.\ “am weitesten südlich gelegenen” flips “furthest south” \textcolor{red}{(major accuracy)}; ... “fünftgrößten Ausländern” instead of “fünftmeisten Ausländer” confuses quantity \textcolor{red}{(major)}; ... “bildeten den Rest der zehn Bezirke” is valid but slightly awkward \textcolor{orange}{(minor fluency)}.
\textcolor{purple}{</think>}
}
\caption{Excerpts from DeepSeek R1 reasoning trace for MT evaluation.}
\label{fig:compact-reasoning}
\end{figure}

In both cases, we observe that the model proceeds iteratively through each sentence and phrase of the source text and translation hypothesis. It starts by identifying errors, then assigns the appropriate category, and finally determines severity. In the effective reasoning example (see excerpts from it in Fig.~\ref{fig:compact-reasoning}), the model correctly identifies all issues. In the ineffective reasoning example, the model identifies two minor issues with wording and structure, yet overestimates their severity.

%% file: structure/7_discussion.tex
\section{Discussion}

Our experimental results across MT evaluation, TS evaluation, and reasoning token counts analysis reveal several noteworthy patterns regarding the utility of reasoning capabilities in NLG evaluation.

\begin{figure*}[htbp]
    \centering
    \includegraphics[width=\linewidth]{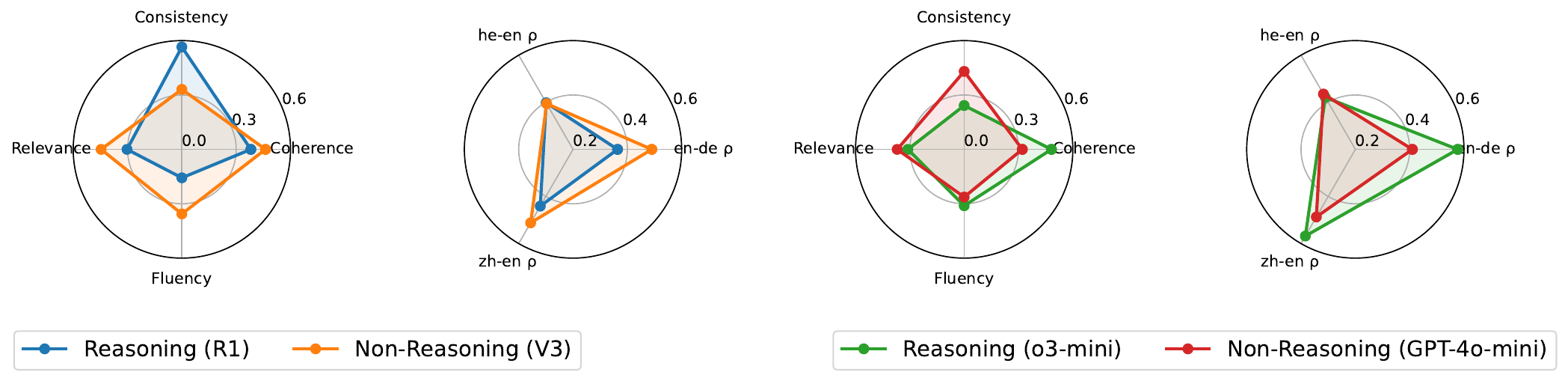}
    \caption{Comparison of evaluation quality of reasoning and non-reasoning LLMs in MT and TS evaluation.}
    \label{fig:radar-chart}
\end{figure*}

\paragraph{Architecture-Specific Performance} The experimental results demonstrate clear performance differences between reasoning model architectures. As shown on Figure~\ref{fig:radar-chart}, DeepSeek-R1 mostly underperforms compared to its non-reasoning counterpart (DeepSeek V3) in both MT and TS evaluation. In contrast, OpenAI o3-mini models demonstrates superior performance over their non-reasoning equivalent (GPT-4o-mini) in several evaluation scenarios, particularly for MT. This architecture-dependent performance demonstrates that reasoning capabilities alone do not guarantee improved evaluation quality, rather the specific implementation and post-training approach for enhancing reasoning capabilities determines effectiveness.

\paragraph{DeepSeek Limitations} DeepSeek-R1 achieves lower correlation with human judgments compared to DeepSeek V3 in both MT (en-de: 0.364 vs.\ 0.490 $\downarrow$25.7\%, zh-en: 0.441 vs.\ 0.512 $\downarrow$13.9\%) and TS tasks (coherence: 0.381 vs.\ 0.462 $\downarrow$17.5\%, relevance: 0.303 vs.\ 0.446 $\downarrow$32.1\%). Insufficient multilingual training data or lack of specific fine-tuning for evaluation tasks likely contribute to this performance gap. While DeepSeek-R1 excels in consistency evaluation for TS (0.565 vs.\ 0.331 $\uparrow$70.7\%), its generally weaker performance indicates that its reasoning approach does not align with the process of fine-grained NLG evaluation.

\paragraph{OpenAI o3-mini Effectiveness} In contrast, OpenAI o3-mini demonstrate superior evaluation capabilities, particularly in machine translation. The o3-mini-high model achieves the highest correlation scores for en-de (0.577) and zh-en (0.568) translation pairs, substantially outperforming the non-reasoning GPT-4o-mini (0.410 and 0.487, $\uparrow$40.7\% and $\uparrow$16.6\% respectively). Our correlation analysis quantifies two distinct relationships. First, o3-mini models show clear negative correlations between reasoning token count and evaluation error (-0.118, -0.110, and -0.091 for high, medium, and low reasoning effort in MT evaluation). This supports the test-time scaling hypothesis: allowing the model to produce more reasoning tokens consistently reduces prediction error for this architecture. Second, we observe strong negative correlation between reasoning token count and model-predicted scores for o3-mini-high (-0.474 for MT, -0.209 for TS). This indicates that models engage in substantially more extensive reasoning when they detect potential issues in the text, regardless of whether these issues align with human judgments. The correlation strengthens with increasing reasoning effort settings, further confirming this pattern of deeper reasoning for perceived lower-quality content.

\paragraph{Distillation Efficacy} Our results reveal variable success in distilling reasoning capabilities for evaluation tasks. R1 Qwen 32B maintains performance comparable to the original DeepSeek-R1 on SummEval (0.355 vs.\ 0.351) and achieves 97\% of its performance on Eval4NLP (0.564 vs.\ 0.583). Similarly, R1 LLaMa 70B maintains or improves performance compared to DeepSeek-R1 in key metrics, notably outperforming the original model on MT evaluation for en-de (0.421 vs.\ 0.364, $\uparrow$15.7\%) and zh-en (0.451 vs.\ 0.441, $\uparrow$2.3\%), and achieving improvements in TS fluency evaluation (0.330 vs.\ 0.157, $\uparrow$110\%) and relevance (0.322 vs.\ 0.303, $\uparrow$6.3\%). In contrast, the smaller R1 LLaMa 8B shows substantial degradation in summarization evaluation, with average correlation dropping from 0.351 to 0.174 (50.4\% reduction). These results demonstrate that effective distillation of reasoning capabilities requires sufficient model capacity, with models of at least 32B parameters maintaining performance while 8B models lose critical evaluation capabilities. 

\paragraph{Task-Specific Reasoning Requirements} The experiment data reveals distinct patterns between tasks. While OpenAI models demonstrate advantages in MT evaluation (40.7\% improvement for en-de), their performance advantage diminishes in summarization tasks (3.2\% reduction in SummEval average compared to GPT-4o-mini). Correlation analysis further illustrates this task difference, with o3-mini models showing stronger efficacy of increased reasoning token count for MT evaluation than for TS evaluation across all reasoning effort settings. MT evaluation appears to benefit more from the particular reasoning approach employed by o3-mini models. 


These findings demonstrate that reasoning capabilities improve evaluation performance very selectively, with efficacy dependent on architectural implementation, task-specific alignment, and reasoning intensity. The superior performance of o3-mini models, particularly in MT evaluation, suggests that they incorporate post-training elements better suited for translation evaluation tasks, potentially including broader multilingual exposure or specific fine-tuning for translation assessment. Future work should investigate targeted tuning to reasoning approaches specifically designed for NLG evaluation contexts, as well as the performance of other reasoning model architectures.

%% file: structure/8_conclusion.tex
\section{Conclusion}

This study presents the first systematic comparison of reasoning versus non-reasoning LLMs as NLG evaluators across MT and TS tasks. Our analysis demonstrates that the impact of reasoning capabilities on evaluation quality varies significantly by model architecture, with distinct efficacy patterns emerging across different evaluation dimensions.

The OpenAI o3-mini model consistently outperforms its non-reasoning counterpart on MT evaluation metrics, improving correlation with human judgments by 40.7\% for en-de translation. However, DeepSeek-R1 and its distilled versions generally underperform compared to their non-reasoning counterparts across most evaluation tasks, with notable exceptions in TS consistency evaluation, where DeepSeek-R1 shows a 70.7\% improvement over DeepSeek V3. These architecture-dependent patterns demonstrate that reasoning capabilities utility for NLG evaluation depends on their specific implementation and integration with model.

Our correlation analysis reveals two important patterns: first, a low-to-medium correlation between reasoning token count and evaluation error (supporting test-time scaling hypothesis) for OpenAI o3-mini on MT evaluation and R1-Distill-Qwen-32B on TS evaluation, with minimal correlation for other models; second, a strong negative correlation between reasoning token usage and model-predicted quality scores across all models, showing that models dedicate substantially more reasoning tokens when they identify quality issues in the evaluated text.

Regarding distillation efficacy, medium-sized models (R1 Qwen 32B) preserve most evaluation capabilities of the original model despite having 21 times fewer parameters. R1 LLaMa 70B shows mixed results across different evaluation dimensions, while smaller 8B parameter models exhibit substantial performance degradation, losing over 50\% of evaluation quality. This confirms that effective distillation of evaluation-relevant reasoning requires sufficient model capacity.

These findings highlight that reasoning capabilities can improve NLG evaluation, but their effectiveness depends on model architecture and implementation. Future work should focus on specifically aligning reasoning approaches with evaluation requirements through fine-tuning and architectural modifications that enhance reasoning application to NLG evaluation.

%% file: structure/X-Appendix.tex
\onecolumn
\section{Appendix: Reasoning Examples}
\label{sec:appendix_reasoning}

This appendix presents full reasoning examples from both effective and ineffective cases of LLM reasoning when evaluating translation quality.

\begin{figure}[htbp]
\small{
\textbf{Source:} Statistics Norway's figures showed that foreigners were scattered all over the city's districts, as Søndre Nordstrand, the borough furthest south in Oslo, had over 14,000 immigrants registered there. Many immigrants also lived in sought-after areas, as Frogner was the district with the fifth most foreigners. After that, Stovner, Bjerke, Grorud, Østensjø and Sagene made up the rest of the ten districts in Oslo with the most residents from another country.

\textbf{Hypothesis:} Statistiken Norwegens Zahlen zeigten, dass Ausländer in den Bezirken der Stadt verstreut waren, da Søndre Nordstrand, der Bezirk weiter südlich in Oslo, dort über 14.000 Einwanderer registriert hatte. Viele Einwanderer lebten auch in begehrten Gebieten, da Frogner der Bezirk mit den fünftgrößten Ausländern war. Danach bildeten Stovner, Bjerke, Grorud, Østensjø und Sagene den Rest der zehn Bezirke Oslos mit den meisten Einwohnern aus einem anderen Land.

\textbf{Diff:} $|$Human(-25) - LLM(-25)$|$ = 0.0

\textbf{Reasoning:}
\textcolor{purple}{<think>}
Okay, let's tackle this translation analysis. First, I need to compare the English source text with the German 
translation provided. The goal is to identify any errors in accuracy, fluency, style, terminology, etc., and 
classify them as critical, major, or minor.

Starting with the first sentence: "Statistics Norway's figures showed that foreigners were scattered all over the 
city's districts..." The German translation begins with "Statistiken Norwegens Zahlen zeigten..." Hmm, "Statistics 
Norway" is the name of the organization, so translating it literally as "Statistiken Norwegens" might be a 
mistranslation. The correct name in German should be "Statistik Norwegen" or "Statistisches Amt Norwegens." This is
a terminology error because it's an incorrect translation of a proper noun. Since it's the name of an official 
body, getting it wrong could confuse readers about the source, so this might be a critical error.

Next part: "Søndre Nordstrand, the borough furthest south in Oslo..." is translated as "Søndre Nordstrand, der 
Bezirk weiter südlich in Oslo..." The original says "furthest south," but "weiter südlich" means "further south." 
That's a mistranslation. "Furthest south" should be "am weitesten südlich gelegenen." This affects accuracy because
it changes the geographical reference, which could mislead about the location's position. This is a major error 
since it disrupts the flow but the main idea is still somewhat clear.

Moving on: "Frogner was the district with the fifth most foreigners." The German translation says "Frogner der 
Bezirk mit den fünftgrößten Ausländern war." "Fünftgrößten" translates to "fifth largest," but the original is 
"fifth most." The correct term here should be "die fünftmeisten Ausländer." Using "größten" (largest) instead of 
"meisten" (most) is a mistranslation. This is a major accuracy error because it changes the meaning from quantity 
to size, which is misleading but the context might still let readers infer the correct meaning.

Then, "made up the rest of the ten districts..." is translated as "bildeten den Rest der zehn Bezirke..." The 
original specifies that these districts are part of the ten with the most residents from another country. The 
German translation says "den Rest der zehn Bezirke," implying that these districts complete the list of ten. 
However, the original mentions that after Frogner (which is fifth), Stovner, Bjerke, etc., make up the rest of the 
ten. The translation might be missing clarity here, but it's not a clear omission. Maybe a minor fluency issue 
because "bildeten den Rest" is a bit awkward but understandable.

Fluency check: The sentence structure in German seems mostly correct, but "da" is used twice as a conjunction, 
which in German can sometimes be repetitive. However, "da" is correct here meaning "since" or "as," so that's fine.
No major fluency issues.

Another point: "registriert hatte" is in past perfect, which might not align with the original's simple past 
"had... registered." But in German, the past perfect is sometimes used where English uses simple past, so this 
might be acceptable. Not a critical error.

So compiling the errors:

Critical: The mistranslation of "Statistics Norway" as "Statistiken Norwegens" is a critical terminology error 
because it misrepresents the organization, leading to potential confusion.

Major: The mistranslation of "furthest south" to "weiter südlich" and "fifth most" to "fünftgrößten" are both 
accuracy issues that affect meaning but don't completely hinder understanding.

Minor: The structure "bildeten den Rest der zehn Bezirke" is a bit awkward but still understandable, so a minor 
fluency/style issue.
\textcolor{purple}{</think>}

\textbf{Error Assessment:}

\textcolor{red}{Critical:}  

terminology/inappropriate for context - "Statistiken Norwegens" (incorrect translation of "Statistics Norway"; should be "Statistik Norwegen" or similar).  

\textcolor{orange}{Major:}  

accuracy/mistranslation - "weiter südlich" (translates to "further south" instead of "furthest south"). 

accuracy/mistranslation - "fünftgrößten Ausländern" (incorrectly translates "fifth most foreigners" as "fifth largest foreigners").  

\textcolor{blue}{Minor:}  

fluency/awkward - "bildeten den Rest der zehn Bezirke" (awkward phrasing for "made up the rest of the ten districts").
}
\label{fig:good-reasoning}
\caption{Effective reasoning example}
\end{figure}

\begin{figure*}[htbp]
\small{
\textbf{Source:} (PERSON2) Because like the way I believe Martin does his <unintelligible/> little translation, so he translates more sentences at once and then picks only the centre one and goes like this for the whole document. So, the context is like in one <unintelligible/> together with sentence we actually want to translate.

\textbf{Hypothesis:} (PERSON2) Weil ich glaube, dass Martin seine <unverständlich/> kleine Übersetzung so macht, dass er mehrere Sätze auf einmal übersetzt und dann nur den mittleren auswählt und so das ganze Dokument durchgeht. Also, der Kontext ist wie in einem <unverständlich/> zusammen mit dem Satz, den wir tatsächlich übersetzen wollen.

\textbf{Diff:} $|$Human(-0) - LLM(-10)$|$ = 25.0

\textbf{Reasoning:}
\textcolor{purple}{<think>}Okay, let's tackle this translation error analysis. The user provided an English source text and its German 
translation, and I need to identify and classify the errors.

Looking at the first part: "the way I believe Martin does his <unintelligible/> little translation" becomes "dass 
Martin seine <unverständlich/> kleine Übersetzung so macht". The word "little" is translated as "kleine", which is 
"small" in German. But in context, "little" might be more about the manner of translation rather than size. Maybe 
"kleine" is a mistranslation here. That could be a major accuracy error.

Next, "he translates more sentences at once" is "dass er mehrere Sätze auf einmal übersetzt". "More" here could 
mean "several" or "more than one". The German "mehrere" means "several", which fits. So that's okay.

Then, "picks only the centre one" becomes "nur den mittleren auswählt". "Centre one" translates to "mittleren", 
which is "middle one". That's accurate. No problem here.

"goes like this for the whole document" is "und so das ganze Dokument durchgeht". "Durchgeht" means "goes through",
which makes sense. Maybe "durchgeht" is a bit literal, but acceptable. No major issue.

In the next sentence: "the context is like in one <unintelligible/> together with sentence we actually want to 
translate" becomes "der Kontext ist wie in einem <unverständlich/> zusammen mit dem Satz, den wir tatsächlich 
übersetzen wollen". The structure here seems correct. "zusammen mit dem Satz" is "together with the sentence", 
which matches. 

Now, checking for additions or omissions. The German translation doesn't seem to add or omit any content except the
handling of <unintelligible/>, which is appropriately translated. 

Fluency-wise, the sentence structure in German reads smoothly. No obvious grammar or spelling errors. The register 
is consistent, as it's a conversational tone matching the source. 

Terminology: Terms like "Übersetzung" for "translation" and "Dokument" for "document" are appropriate. No 
inconsistencies here.

Style: The phrase "kleine Übersetzung" might be slightly awkward if "little" was meant to be diminutive or 
pejorative, but in German, "kleine" is neutral. However, if "little" in English is meant to imply a specific 
method, "kleine" might not capture that nuance. This could be a minor style issue.

Wait, the user's example had "involvement" as a mistranslation. Here, "little" to "kleine" might be a 
mistranslation if the intended meaning was different. For example, if "little" refers to a minor or specific part 
of the translation process, "kleine" (small) might not be accurate. That's a possible accuracy/mistranslation, 
major error.

Also, "den wir tatsächlich übersetzen wollen" uses "wollen" (want to), which matches the source's "want to 
translate". That's accurate.

So final classification:

Critical: no-error

Major: accuracy/mistranslation - "kleine"

Minor: None, unless "durchgeht" is considered awkward. But "durchgeht" is correct. Maybe the structure "so macht, 
dass er..." is a bit awkward, leading to a minor fluency/grammar issue. Alternatively, style/awkward.

Alternatively, the phrase "so das ganze Dokument durchgeht" might be better as "so durch das ganze Dokument geht", 
but "durchgeht" is a separable verb. "Durchgeht" is correct, but the word order might be slightly off. Hmm, in 
German, "und so das ganze Dokument durchgeht" is correct. The verb "durchgeht" is at the end, which is standard in 
subordinate clauses. Wait, the structure here is "dass er... übersetzt und dann... auswählt und so... durchgeht". 
The "und so das ganze Dokument durchgeht" is a main clause, so the verb should be in second position. But here, 
it's "und so das ganze Dokument durchgeht" – "durchgeht" is at the end, which is correct for a subordinate clause, 
but after "und", it's a main clause. So maybe the word order is incorrect. That would be a grammar error. Let me 
check:

Original structure: "dass er mehrere Sätze auf einmal übersetzt und dann nur den mittleren auswählt und so das 
ganze Dokument durchgeht."

The "und so das ganze Dokument durchgeht" is connected with "und", so it's a main clause. In main clauses, the verb
should be in second position. So "und so geht er das ganze Dokument durch" would be correct. But the translation 
uses "durchgeht" at the end, which is incorrect. So this is a grammar error. That's a major fluency/grammar error 
because it disrupts the flow. But does it hinder comprehension? Maybe not critically, but it's a clear grammar 
mistake. So major or minor?

In the example, "wäre" was a minor grammar error. Here, incorrect verb placement is a more significant grammar 
issue. So maybe major fluency/grammar.

So adding that:

Major:
accuracy/mistranslation - "kleine"
fluency/grammar - "durchgeht" word order

But wait, the user's example had "wäre" as minor. Maybe this is a major error because it's more disruptive. 
Alternatively, if the sentence is still understandable, it's major. Let's classify it as major.
\textcolor{purple}{</think>}

\textbf{Error Assessment:}

\textcolor{red}{Critical:}  
no-error  

\textcolor{orange}{Major:}  

accuracy/mistranslation - "kleine" (misinterprets "little" as "small" rather than contextual nuance, e.g., "specific" or "particular" method)  

fluency/grammar - "durchgeht" (incorrect verb placement; should follow main clause structure, e.g., "geht ... durch")  

\textcolor{blue}{Minor:}  
no-error
}
\label{fig:bad-reasoning}
\caption{Ineffective Reasoning Example}
\end{figure*}
\newpage
\section{Additional Information}
\subsection{Potential Risks}
Due to reasoning LLMs being a very recent development, little research exists regarding the safety of the reasoning process. Therefore, it is possible (though not observed in our experiment) that reasoning traces of LLMs used in prompt-based evaluation might show certain evidence of misalignment or bias. This could influence generated quality assessments. Further research is necessary to identify and mitigate such risks in reasoning LLMs.

\subsection{Licensing and Intended Use}
We use parts of GEMBA-MQM evaluation code~\footnote{\url{https://github.com/MicrosoftTranslator/GEMBA}}. This code is distributed under CC-BY-SA-4.0 License. Therefore, due to ShareAlike clause, our code is licensed under the same license.

Our use of those artifacts and LLMs that we evaluate is fully consistent with applicable Intended Use policies.

\subsection{PII in data}
We have not performed rigorous checks for PII for any of the datasets that we use. We believe that this is the responsibility of the authors of those datasets. At the same time, we note that WMT23 dataset does indeed protect PII by replacing names with special tags, which can be seen in our examples in Appendix~\ref{sec:appendix_reasoning}.

\subsection{Packages}
We use `mt-metrics-eval' package, version 2, commit `6d4b0bb'. We also use `scipy' in TS meta-evaluation. The version of `scipy' is 1.15.